# Reconstruction of Complex-Valued Fractional Brownian Motion Fields Based on Compressive Sampling and Its Application to PSF Interpolation in Weak Lensing Survey

Andriyan B. Suksmono

*Abstract*— **A new reconstruction method of complex-valued fractional Brownian motion (CV-fBm) field based on Compressive Sampling (CS) is proposed. The decay property of Fourier coefficients magnitude of the fBm signals/fields indicates that fBms are compressible. Therefore, a few numbers of samples will be sufficient for a CS based method to reconstruct the full field. The effectiveness of the proposed method is showed by simulating, random sampling, and reconstructing CV-fBm fields. Performance evaluation shows advantages of the proposed method over boxcar filtering and thin plate methods. It is also found that the reconstruction performance depends on both of the fBm's Hurst parameter and the number of samples, which in fact is consistent with the CS reconstruction theory. In contrast to other fBm or fractal interpolation methods, the proposed CS based method does not require the knowledge of fractal parameters in the reconstruction process; the inherent sparsity is just sufficient for the CS to do the reconstruction. Potential applicability of the proposed method in weak gravitational lensing survey, particularly for interpolating non-smooth PSF (Point Spread Function) distribution representing distortion by a turbulent field is also discussed.**

# I. INTRODUCTION

Two-dimensional fBm (fractional Brownian motion) is a well-suited model of various natural objects; such as clouds, terrain elevation, interstellar medium, and galactic star fields [1]-[3]. The fBm is also called $1/f$ noise, which is attributed to the exponential decay of the magnitude spectrum over frequency. Suitability of the fBm to model terrain elevation has been shown in [4], where $1/f$ based frequency synthesis scheme is employed to reduce distortion in phase unwrapping of InSAR (Interferometric Synthetic Aperture Radar) image.

Furthermore, some kinds of images are appropriately modeled by complex-valued field. As an example, the magnitude and phase image of the InSAR can be combined to form a single complex-valued image. Algorithms exploring the complex-valued nature of the image will capable to perform better than the ones treating the images separately [5]-[7]. Ellipticity information of PSF (Point Spread Function) and shear gravitational field causing distortion on galaxy images can also be represented as a complex-valued image. In the weak (-gravitational) lensing survey, interpolation of PSF values at particular spatial position, given sorrounding PSF values is an important issue for accurate measurement and mapping of the darkmatter field [8]-[10].

In this paper, a new method to reconstruct or interpolate complex-valued fBm field (CV-fBm) from a few number of samples taken at random positions is presented. The method is based on the fact that the magnitude of Fourier coefficients in the fBm follows power law decay [11], indicating that fBm is a compressible field and therefore can be well-approximated by a *K*-sparse signal in frequency domain. By using an emerging Compressive Sampling (CS) method [12]-[15], the interpolation problem can be solved by a CS reconstruction algorithm. In the previous works, CS has been successfully applied in VLBI (Very Long Baseline Interferometry) [16] and GPR (Ground Penetrating Radar) imaging [17].

Applicability of the proposed method to solve PSF reconstruction problem is explored. Weak-(gravitational) lensing survey needs the PSF values at particular position coincides with the galaxy image, whereas only a few number of the PSF values derived from star image at other places are known. In particular, the difficult case of turbulent field induced by atmospheric disturbance will be considered [8]-[10], and CV-fBm is assumed as a model of such a field. The proposed CS reconstruction will be compared with smooth interpolation methods, where RMSE (Root Mean Squared Error) is used to measure the performance.

The rest of the paper is organized as follows. Section II describes briefly the CV-fBm fields and representation of PSF distribution by the CV-fBm. Section III described boxcar filtering, thin plate method, and formulation of the proposed CS reconstruction method. Experiments and analysis with both of simulated CV-fBm and PSF distribution will be given in Section IV and Section V concludes the paper.

## II. CV-fBM Field and PSF Distribution

### A. Formulation and Synthesis of CV- fBm Fields

A two-dimensional fBm is a nonstationary random field with statistically self-similar property determined by Hurst exponent $0<H<1$. In the frequency domain, the fBm is characterized by $1/f$ property, which means that, the magnitude of the (spatial-) frequency domain of an fBm follows the power-law decay given by [11]:

$$S_H(\omega) \sim 1/|\omega|^{2H+1} \tag{1}$$

where $\omega = \sqrt{\omega_x^2 + \omega_y^2}$, with $\omega_x$ and $\omega_y$ are spatial frequency in $x$ and $y$ directions, respectively. When the field values in the spatial domain are complex, it will be called a complex-valued fBm (CV-fBm) field.

The CV-fBm field can be constructed by Fourier synthesis as follows [18]. First, complex-valued white noise $s$ is generated on a $2M \times 2N$ grid. Then, it is Fourier transformed into $S$ and the magnitude coefficient is adjusted to follow (1) which yield $S_H$. The CV-fBm field $\breve{s}_H$ is obtained by inverse-Fourier transforming $S_H$. Then, four $M \times N$ sub-images are obtained, each is a mirror image of the other, therefore it is sufficient to take only one of them as the generated CV-fBm image $s_H$. Fig.1 shows synthesize CV-fBm image on a grid of $100 \times 100$ size with $H=0.8$, displayed as a pair of (a) real- and (b) imaginary- images.

### B. Representation of PSF Distribution by CV-fBm Field

The real-imaginary pair of the CV-fBm field can be used to represent an actual physical property. In a weak lensing survey, intensity distribution of a PSF is derived from a star image. By denoting the ellipticity components as $e_1$ and $e_2$, ellipticity of the PSF of an image $I_p \equiv I(x_p, y_p)$ is obtained from originally circular's profile $I_o(x_o, y_o)$ by the following shear transform

$$\begin{pmatrix} x_p \\ y_p \end{pmatrix} = \begin{pmatrix} 1-e_1 & -e_2 \\ -e_2 & 1+e_1 \end{pmatrix} \begin{pmatrix} x_0 \\ y_0 \end{pmatrix} \qquad (2)$$

Furthermore, the two-component PSF ellipticities can be expressed as a single complex-valued quantity

$$e_{PSF} = e_1 + ie_2 \qquad (3)$$

where $i = \sqrt{-1}$ is the imaginary number.

Ellipticity of the PSF are originated from optical distortion caused by imperfection of the telescope or due to atmospheric disturbance, which are spatially varying functions. Therefore, the complex-representation of the PSF ellipticities that is distributed spatially is actually a complex-valued random field (CVRF).

The PSF distribution can be divided into two categories, i.e. slowly varying and fast-varying fields. For the slowly varying case, PSF interpolation can be conducted relatively easily by approximation with smooth functions. The fast varying case is a more challenging problem because the smooth functions will not sufficient to approximate the irregularity of the PSF field. In this paper, turbulence induced fast varying PSF field will be modeled as a CV-fBm, so that the ellipticity distribution of the field is like ones displayed in Fig.1.

In the Quadrature Moment (QM) framework [8], the ellipticity of a star image $I_p$ is calculated (repeated here for clarity) as

$$e_{PSF} = \frac{q_{11}^2 - q_{22}^2 + 2i\, q_{12}}{q_{11}^2 + q_{22}^2 + 2(q_{11}q_{22} - q_{12}^2)^{1/2}} \qquad (4)$$

with $q_{ij}$ denotes second order brightness moment defined as

$$q_{ij} = \frac{\sum_p w_p I_p (\theta_i - \hat{\theta}_i)(\theta_j - \hat{\theta}_j)}{\sum_p w_p I_p} \tag{5}$$

In equation (5), $I_p$ is the intensity or flux of the photon of the $p^{th}$ pixel, $w_p$ the weight function, and $\theta_i$ is the pixel position in row-column index or $x$ and $y$ coordinates, i.e. $\theta_1 = x_p$ and $\theta_2 = y_p$. Then, the PSF radius $R_{PSF}$ is calculated as

$$R_{PSF} = \sqrt{q_{11} + q_{22}} \tag{6}$$

In the real case, parametric representation of the PSF in term of ellipticity and radius or FWHM (Full Width Half Maximum) is calculated from (noisy) star images observed by a telescope. By using Eqs. (4), (5), and (6) for each star image, distribution of the parameters are obtained; each one is essentially a random field, with (4) represents a CVRF.

## III. RECONSTRUCTION OF PSF FIELD FROM SPARSE SAMPLES

In this section, three reconstruction methods of spatially distributed PSF represented as complex-valued fields, i.e., the (complex-valued) boxcar filter, thin-plate, and the proposed CS methods, are described. The PSF interpolation problem deals with estimation of PSF values at particular position, given only a small number of PSF values at other places. Assuming that there is an underlying PSF distribution model, the unknown PSF values is calculated as follows:

a) Convert known star images $I_p$ into parameters: i.e. complex ellipticities showed in (4) and star radius or FWHM in (5) under particular intensity distribution, such as Airy or Moffat. A set of ellipticities values at various positions $e_{\text{PSF}}$ are obtained, which is a complex-valued field, and radius distribution $R$ of real-valued field. Both $e_{\text{PSF}}$ and $R_{\text{PSF}}$ only known at a few number of (discretized) subsample positions, denoted by $e_{PSF}^{sub}$ and $R_{PSF}^{sub}$.

b) Interpolate the known PSF parameters at asked position: i.e, calculate $\hat{e}_{PSF}$ and $\hat{R}_{PSF}$ at desired positions.

c) Optionally, convert $\hat{e}_{PSF}$ and $\hat{R}_{PSF}$ back into estimated star images $\hat{I}_p$ that represents 2D intensity distribution of the PSF.

The performance of interpolation algorithm can be measured by computing the mean squared-error of both the ellipticity

$$\sigma^2(e_{PSF}) = \left\langle (\hat{e}_{PSF} - e_{PSF})^2 \right\rangle \tag{7}$$

and the radius

$$\sigma^2(R_{PSF}) = \left\langle (\hat{R}_{PSF} - R_{PSF})^2 \right\rangle \tag{8}$$

where $e_{PSF}$ dan $R_{PSF}$ are the true values of ellipticity and radius, respectively, and $\langle...\rangle$ means taking expectation value or averaging.

This paper will concentrate to measure the performance of the proposed compressive sampling method, compared to boxcar filter and thin plate method. In particular, the main interest is to calculate ellipticity distortion of the methods in term of RMSE (Root Mean-Square-Error) derived from (7),

$$RMSE(e_{PSF}) = \sqrt{\langle(\hat{e}_{PSF} - e_{PSF})^2\rangle} \qquad (9)$$

*A. Boxcar Filtering Method*

The boxcar filter or sliding window method is the simplest among the three, which essentially is averaging. First, a window of size $W \times W$ is selected. Starting from a particular corner, e.g. left uppermost, the center of the window is located at a point in the PSF grid, then the value is replaced by the average value of all of the known PSF values inside the window. This process is repeated to all the remaining point in the grid. At the end of the process, dynamic range of the result is adjusted with the known values from sparse data sample as a reference.

*B. Thin Plate Method*

The thin plate method that is used in this paper is adopted from Matlab's thin-plate smoothing spline function *tpaps*. This method interpolates the data values given by the array $e_{PSF}^{sub}$ for a given positions listed in array $x^{sub}$ by minimizing weighted sum

$$\min\ pE(\hat{e}_{PSF}) + (1-p)R(\hat{e}_{PSF}) \qquad (10)$$

with $E(\hat{e}_{PSF})$ an error measure defined by

$$E(\hat{e}_{PSF}) = \sum \left| e_{PSF}^{sub} - \hat{e}_{PSF}(x^{sub}) \right|^2 \qquad (11)$$

and roughness $R(\hat{e}_{PSF})$ is given by

$$R(\hat{e}_{PSF}) = \int \left( |D_1 D_1 \hat{e}_{PSF}|^2 + 2|D_1 D_2 \hat{e}_{PSF}|^2 + |D_2 D_2 \hat{e}_{PSF}|^2 \right) \quad (12)$$

where $D_i f$ denotes the partial derivative with respect to the $i^{th}$ argument. Detail description of the function can be found in Matlab's Curve Fitting Toolbox User Guide [19].

*C. Compressive Sampling Method*

The CS is an emerging paradigm that unifies sampling and compression [12]-[14]. When a signal is sparse on a particular bases Ψ, while the measurement is performed by a projection bases Φ, the coherence between these bases is defined as

$$\mu(\Phi, \Psi) = \sqrt{N} \max_{1 \leq k, j \leq N} \left| \langle \phi_k, \psi_j \rangle \right| \quad (13)$$

According to the CS theory, a few number *M* of the subsamples

$$M \geq C \cdot \mu^2(\Phi, \Psi) \cdot K \cdot log(N) \quad (14)$$

is sufficient to reconstruct fully the original signal. In (14), *C* is a positive constant, whereas *K* denotes the sparsity of the signal.

Most of real-life signals are only nearly sparse or also called compressible. Consider a signal $e_{PSF}$ expressed in a sparsity basis Ψ, i.e, $e_{PSF} = \Psi a$, which is measured through a projection by Φ that yield subsamples $e_{PSF}^{sub} = \Phi \Psi a$. Generally, CS recovery seeks for a set of coefficient *a* that minimize $L_1$-norm of the solution. The recovery yields an approximation of $e_{PSF}$, denoted by $\hat{e}_{PSF}$, by *K* number of coefficients $a_k = \{a_1, a_2, ..., a_K\}$. A signal is compressible if the sorted magnitude coefficients $|a_k^s|$ decay rapidly, i.e.,

$$|a_k^s| \leq C_1 \cdot k^{-q}, \quad k = 1, 2, ..., K \tag{15}$$

where $C_1$ and $q$ are nonnegative constants. The upper bound of the error is given by [15]

$$\|e_{PSF} - \hat{e}_{PSF}\| \leq C_2 \cdot K^{1/2-q} \tag{16}$$

where $C_2$ is a nonnegative constant.

Equation (1) indicates that the magnitude of the Fourier coefficients of the fBm field decays as the power of the frequency. The speed of the decay is determined by the value of the Hurst parameter; which means that the decay will be faster for a larger value of $H$. Therefore, according to (15), the frequency domain representation indicates compressibility of the fBm field. The fBm can be considered as a nearly sparse field in the Fourier domain, whose degree of sparsity is determined by the Hurst parameter.

*3.1 CS Reconstruction by Basis Pursuit Algorithm*

In the BP (Basis Pursuit) reconstruction scheme, the solution $\hat{e}_{PSF}$ is chosen so that the $L_1$ norm is minimized. To be more general, a complex-valued $N$-length (where $N = R \times C$ for a two-dimensional signal of $R$-rows and $N$-column) discrete fBm signal $e_{PSF}$ is considered, whose subsample is denoted by $e_{PSF}^{sub}$. Since the fBm is sparse in frequency domain, the discrete Fourier transform is chosen to be the sparsity bases $\Psi$.

In the sparse reconstruction problem, the random subsamples (and their positions) are known. The projection basis $\Phi$ is built from the location of known values $e_{PSF}^{sub}$ by constructing a zero-one $M \times N$ random matrix whose entries is set so that $\Phi e_{PSF} = e_{PSF}^{sub}$. Then, a matrix $A$ is formed by multiplying the projection bases $\Phi$ with the DFT bases $\Psi$, i.e, $A=\Phi\Psi$. Considering Fourier relationship $e_{PSF} = \Psi E_{PSF}$ where $E_{PSF}$ is a vector of Fourier coefficients, the minimization is now given by

$$\min \|E_{PSF}\|_{l_1} \quad subject \quad to \quad A E_{PSF} = e_{PSF}^{sub} \tag{17}$$

Optimization in the equation can be solved by convex programming [20]. The optimization process yield optimum solution $E_{PSF}^*$, which are Fourier coefficients of the fBm signal. The reconstruction of the spatial domain fBm field is conducted by Fourier transform

$$\hat{e}_{PSF} = \Psi E_{PSF}^* \tag{18}$$

*3.2. CS Reconstruction by TV Minimization*

Total Variance (TV) minimization method has been employed to solve the problem of reconstruction from partial Fourier samples (RPFS). In [13], a phantom image simulating MRI imaging was successfully recovered from a few number of radial Fourier coefficients by using TV. Similar case has also been shown in CS-VLBI [16] for a 2D problem and CS-GPR [17] for the 1D case. The fBm reconstruction from sparse random samples can also be formulated similarly, with minor modification.

In the RFPS problems, one wants to reconstruct a space/time signal $g$ from a few number of random Fourier coefficients of the signals $G^{sub}$. The TV optimization assumes that $g$ is a TV-sparse or a smooth function. The partial Fourier operator $B$ is constructed based on the position of known subsample, which is almost similar to the previous BP case. Then, the following TV optimization is conducted

$$\min TV(g) \quad subject \quad to \quad B g = G^{sub} \tag{19}$$

Considering 1/$f$ spectrum of the fBm given by (1), the fBm can be considered as a smooth function in the transform domain. Sparse samples in spatial domain $e_{PSF}$ is reconstructed by seeking the most TV-sparse Fourier transform signal $E_{PSF}$ as follows

$$\min TV(E_{PSF}) \quad subject \quad to \quad B^{-1} E_{PSF} = e_{PSF}^{sub} \tag{20}$$

where $B^{-1}$ is the inverse of partial Fourier transform. After finding the transform domain TV-sparse solution $E_{PSF}^*$, the fBm field is obtained by taking its inverse Fourier transform as in (18).

*3.3 CV-fBm Field Reconstruction by Using TwIST*

Typically, the size the grid in the PSF reconstruction is rather large. For 100×100 size of grid, such as represented by CV-fBm in Fig.1, a faster algorithm is required in the implementation. The TwIST (Two-step Iterative Shrinkage/ Tresholding) algorithm [21] has been used. In short, the TwIST solve linear inverse problem by searching for the minimizer of the objective function

$$\varepsilon(x) = \frac{1}{2}\|y - Ax\|^2 + \lambda \Theta(x) \tag{21}$$

where $\Theta(x)$ is a regularizer. Especially, the fBm reconstruction problem uses TV as the regularizer, therefore, $\Theta(x)=TV(x)$. By adjusting the notations to the PSF estimation case, the objective function becomes

$$\varepsilon(\breve{E}_{PSF}) = \frac{1}{2}\|\breve{e}_{PSF}^{sub} - A \cdot \breve{E}_{PSF}\|^2 + \lambda \cdot TV(\breve{E}_{PSF}) \tag{22}$$

Since reconstructed CV-fBm field, as mentioned before, is obtained by inverse Fourier transforming the transform-domain solution, the observed 2D data should first be prepared to conform to periodic boundary condition. The MATLAB implementation of TwIST takes *y*, *A*, *λ* (and a few other optional parameters) as the inputs. Then, the processing stage for a known 2D spatially distributed complex-ellipticity subsamples $e_{PSF}^{sub}$ located at 2D discrete grid, $x_{PSF}^{sub}$ can be formulated as in Algorithm 1.

## IV. EXPERIMENTS AND ANALYSIS

### A. Simulation of CV-fBm Field

The value of *H* determines the statistical property and the appearance of an fBm field. In the spatial domain, lower value of *H* corresponds to a rough field, whereas the higher ones correspond to a smoother field. In frequency domain, *H* determines the decay speed of the Fourier coefficient. High value of *H* makes the coefficient rapidly decaying, whereas the low values makes the coefficient slowly decaying. Fast decays makes the spectrum more concentrated around zero frequency, whereas slow decay makes the coefficients more spread over the frequency domain.

Figure 2 shows fBm field with *H*=0.2, where the real part is on the left and the imaginary part is on the right. Compare to the fBm with *H*=0.8 in Fig.1, these images have a higher degree of roughness. The spectrum for both of the fBms with *H*=0.2 and *H*=0.8 are shown in Fig.3. The figures show that the spectrum is more contrentated for higher *H* values than the lower one. This fact indicates that fBm with high *H* is more compressible then the one with lower *H*.

### B. Sparse Sampling and Reconstruction of CV-fBm Field

The $1/f$ property indicates that the sparsity of an fBm field is related to Hurst parameter, i.e., the sparsity is high for large *H* and low for small *H*. Since the minimum required number for exact reconstruction *M* is proportional to the sparsity *K*, it can be predicted that for a given number of *M*, the SNR for reconstructed fBm fields with large *H* will be higher than the smaller one.

Paper [22] shows such a case for one dimensional fBm. The similar experiment for the 2D complex-valued fBm fields with 64×64 size has been conducted. Although both of the BP and TV method can be used for sparse reconstruction of fBm field, the existing package developed by CS communities favors TV for computational efficiency. Especially, for a given limited resource of the present days PC capability, BP reconstruction for size larger than 64×64 is computationally impractical.

Table 1 shows the average SNR of ten times repeated reconstruction experiments with L1-magic [23] for various values of *H* and subsampling factor. The result justify that the sparsity of the fBm field is determined by the Hurst parameter *H*.

*C. Reconstruction of Fast-Varying PSF Field*

In the last experiment, performance of the proposed method to interpolate the PSF field will be demonstrated. Six CV-fBm fields of 100×100 size representing ellipticity values with *H* ranging from 0.4 to 0.9 are generated, and the dynamic range has been adjusted for typical ellipticity range of values similar to the GREAT10 database described in [8]. Three sampling cases are considered, i.e. 500, 1000, and 2000 subsamples out of 10,000 total ellipticity values are randomly taken. The performance of the proposed CS algorithm will be compared with boxcar and thin-plate methods. Images displayed in this section are experiment results for interpolating a number of 2000 subsamples of CV-fBm image with *H*=0.8 that represents ellipticity distribution of the PSF showed in Fig.1, where real part in (a) represent $e_1$ and imaginary part in (b) corresponds to $e_2$. The CS reconstruction is based on Algorithm 1 employing TwIST.

Fig.4 displays the solution of the boxcar method, whereas Fig.5 is thin-plate solution, and Fig.6 is obtained by the proposed CS method. The first result shown in Fig.4 indicates that 2% of subsamples is not sufficient for the boxcar method to interpolate the data correctly.

Solution given in Fig.5 and Fig.6 shows distinctive feature of thin-plate and CS methods. Whereas the thin-plate give smooth solution, the CS yields non-smooth one. Measurement of RMSE for both methods shows that the CS gives a better result.

Table 2 shows RMSE performance of the boxcar filter (BOX), thin-plate (TP), and Compressive Sampling (CS) methods, for subsamples number (*NSUB*) 500, 1000, and 2000. The Hurst parameter *H* is varied from 0.9 down to 0.4. The first row with *H*=1.* indicates that the data is taken from file-1, set-1 of GREAT10 database as a reference; which is a smooth function. The table shows that in general boxcar method performs worst among the three, with MSE behavior is almost at random. Nevertheless, it is noticed that for 0.9<*H*<05, the number of subsamples correlate positively with the performance, i.e., interpolation with high sample number performs better than the smaller one.

Comparing the TP and CS method for 0.9<*H*<0.6, the CS method consistently outperform TP on all size of subsample number. Both of them perform better for higher number of subsample. When the subsample is increased to 2000, CS outperform both of boxcar and TP for 0.9<*H*<0.4. This fact is consistent with the CS theory, especially to the requirement of minimum sample number expressed in Eq. (14).

Figure 7 shows the reconstruction results displayed partially as a one-dimensional signal for the sample number 101 to 200. The curves display the result of each method to reconstruct CV-fBm (*H=0.8*) with *NSUB*=2000. It is observed that, the CS solution capable to follow irregularity of the CV-fBm samples, whereas thin-plate only gives smooth approximation. Although the boxcar method also provides irregular solution, the number of subsamples is too low to interpolate properly, which yields a poor performance.

## V. CONCLUSIONS AND FURTHER DIRECTIONS

A new reconstruction method for CV-fBm fields based on sparse samples has been presented. Experiments with simulated image show the effectiveness of the proposed method. The proposed method also has been demonstrated for PSF field interpolation. Comparison with smooth interpolation methods, i.e. the boxcar and thin plate methods, shows that the method perform better; especially for $H$ values ranging from 0.9 to 0.6 for low subsample number, and down to $H$=0.4 for higher subsample number.

List of Figures

**Fig.1** Synthesized CV-fBm field with *H*=0.8: (a) real-part, (b) imaginary part

**Fig.2** The CV- fBm fields with *H*=0.2: (a) real and (b) imaginary-part

**Fig.3** Spectrum of fBm images with different values of *H*: (a) *H*=0.2 (b) *H*=0.8

**Fig.4** Estimated CV-fBm images by boxcar filter with window size 11×11

**Fig.5** Estimated CV-fBm images by thin-plate method

**Fig.6** Estimated CV-fBm images by CS method

**Fig.7** Comparison of reconstruction result as one-dimensional signal by the boxcar, TP and CS methods. Top is real-part, bottom is the imaginary.

List of Tables

List of Algorithms

**Algorithm 1.** Compressive Sampling Reconstruction of CV-fBm field by using TwIST

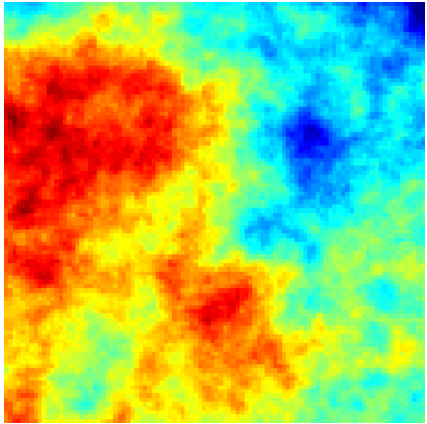 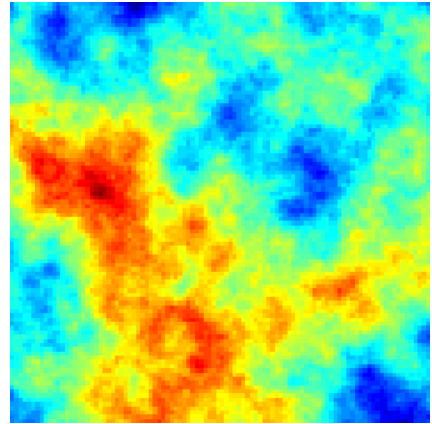

(a) (b)

**Fig.1** Synthesized CV-fBm field with *H*=0.8: (a) real-part, (b) imaginary part

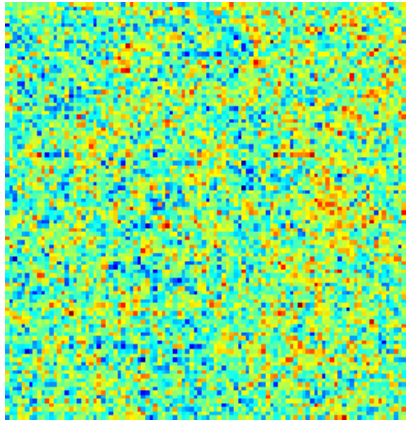 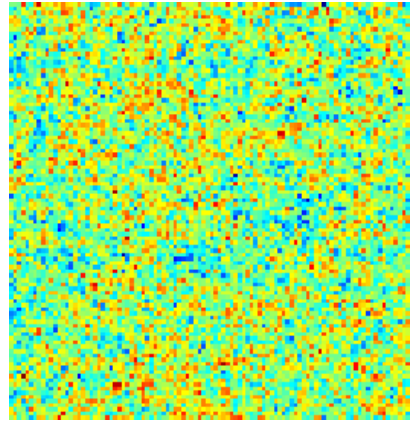

(a) (b)

**Fig.2** The CV- fBm fields with *H*=0.2: (a) real and (b) imaginary-part

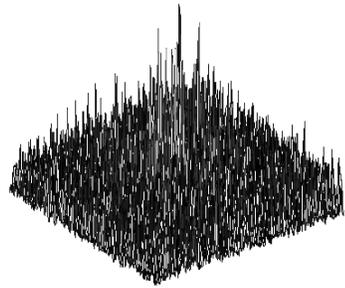
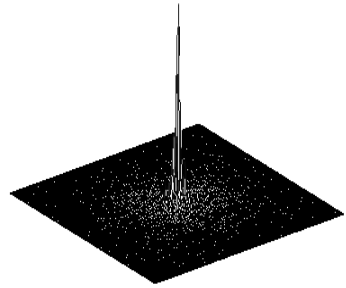

(a)                  (b)

**Fig.3** Spectrum of fBm images with different values of *H*: (a) *H*=0.2 (b) *H*=0.8

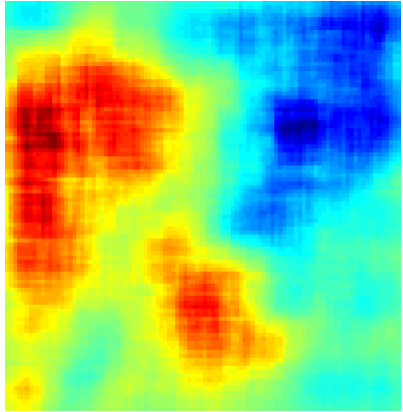 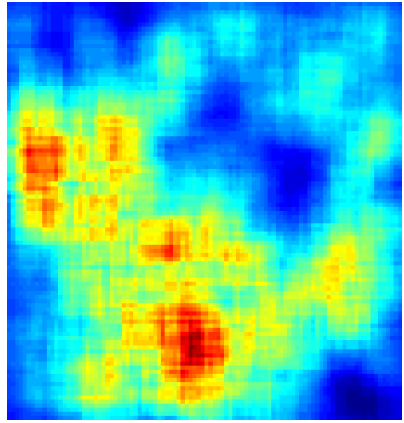

(a) (b)

**Fig.4** Estimated CV-fBm images by boxcar filter with window size 11×11

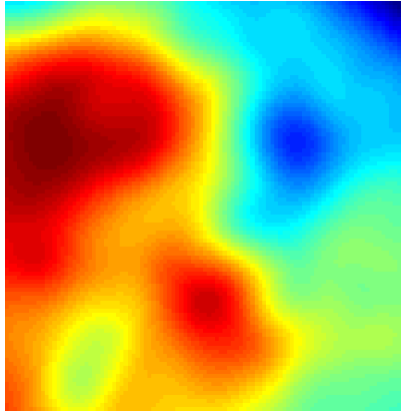 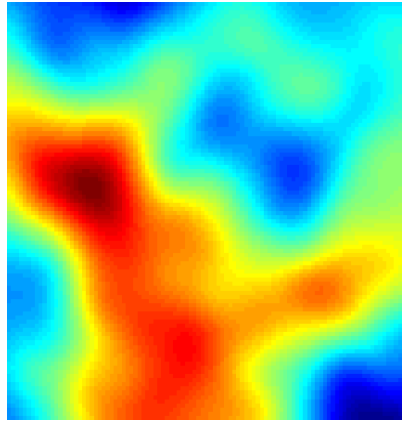

(a)　　　　　　　　　　　　　　　　　　　(b)

**Fig.5** Estimated CV-fBm images by thin-plate method

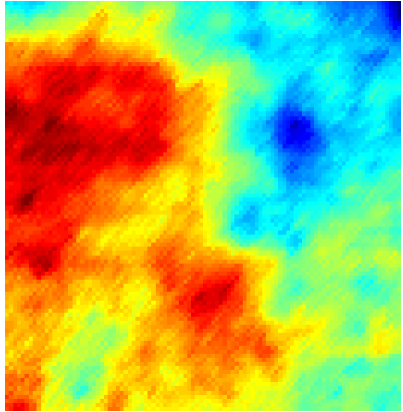
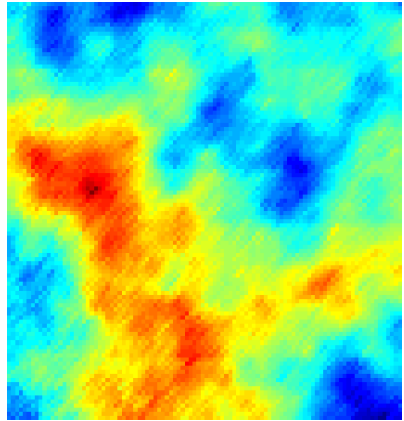

(a)                                      (b)

**Fig.6** Estimated CV-fBm images by CS method

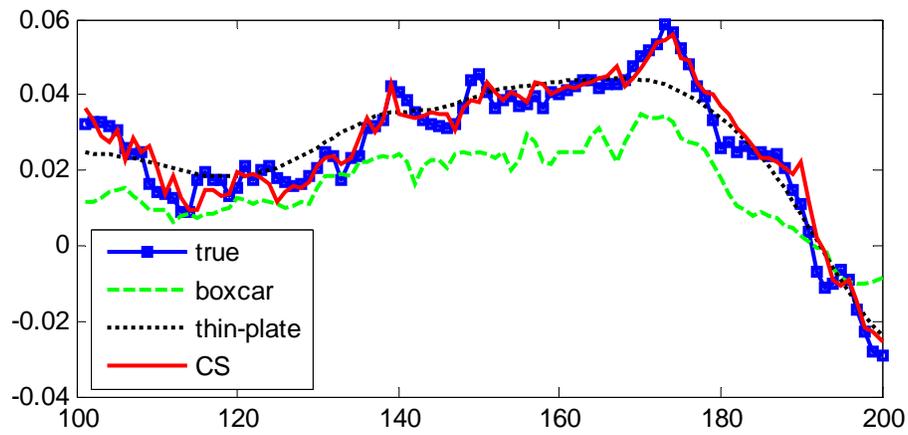

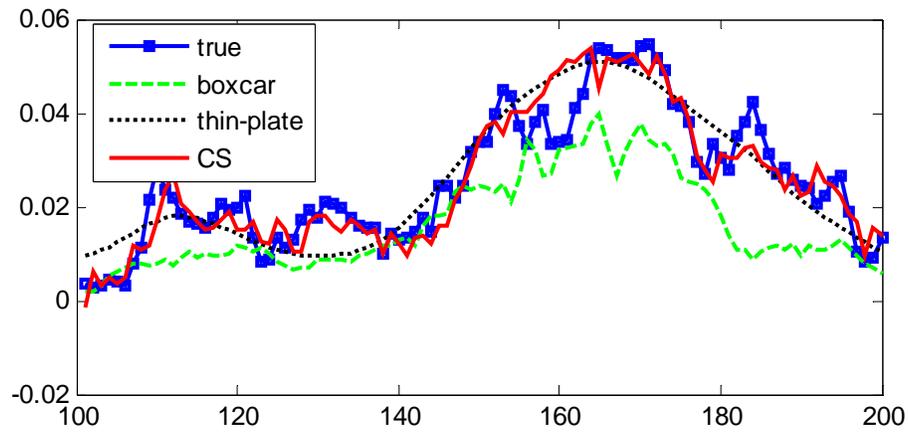

**Fig.7** Comparison of reconstruction result as one-dimensional signal by the boxcar, TP and CS methods. Top is real-part, bottom is the imaginary.

| No | Subsampling Factor | $H$ | Average SNR CS-TV (dB) |
|---|---|---|---|
| 1 | 2 | 0.8 | 16.2 |
| 2 | 2 | 0.6 | 8.4 |
| 3 | 2 | 0.4 | 2.4 |
| 4 | 4 | 0.8 | 13.2 |
| 5 | 4 | 0.6 | 6.1 |
| 6 | 4 | 0.4 | 1.1 |

**Table 1.** Reconstruction Quality of Sparse fBm

| NSUB | H | RMSE | | |
|---|---|---|---|---|
| | | BOX | TP | CS |
| 500 | 1.* | $2.44 \times 10^{-2}$ | $1.21 \times 10^{-3}$ | $1.46 \times 10^{-3}$ |
| | 0.9 | $1.95 \times 10^{-2}$ | $5.86 \times 10^{-3}$ | $4.46 \times 10^{-3}$ |
| | 0.8 | $1.74 \times 10^{-2}$ | $7.65 \times 10^{-3}$ | $6.47 \times 10^{-3}$ |
| | 0.7 | $1.58 \times 10^{-2}$ | $1.10 \times 10^{-2}$ | $9.92 \times 10^{-3}$ |
| | 0.6 | $1.63 \times 10^{-2}$ | $1.50 \times 10^{-2}$ | $1.49 \times 10^{-2}$ |
| | 0.5 | $1.87 \times 10^{-2}$ | $1.84 \times 10^{-2}$ | $1.91 \times 10^{-2}$ |
| | 0.4 | $2.24 \times 10^{-2}$ | $2.07 \times 10^{-2}$ | $2.22 \times 10^{-2}$ |
| 1000 | 1.* | $2.04 \times 10^{-2}$ | $1.13 \times 10^{-3}$ | $1.14 \times 10^{-3}$ |
| | 0.9 | $1.40 \times 10^{-2}$ | $4.99 \times 10^{-3}$ | $3.41 \times 10^{-3}$ |
| | 0.8 | $1.44 \times 10^{-2}$ | $6.88 \times 10^{-3}$ | $5.36 \times 10^{-3}$ |
| | 0.7 | $1.48 \times 10^{-2}$ | $9.94 \times 10^{-3}$ | $8.31 \times 10^{-3}$ |
| | 0.6 | $1.45 \times 10^{-2}$ | $1.39 \times 10^{-2}$ | $1.31 \times 10^{-2}$ |
| | 0.5 | $1.82 \times 10^{-2}$ | $1.77 \times 10^{-2}$ | $1.77 \times 10^{-2}$ |
| | 0.4 | $2.10 \times 10^{-2}$ | $2.01 \times 10^{-2}$ | $2.08 \times 10^{-2}$ |

**Table 2**. Performance comparison of Boxcar (BOX), Thin-Plate (TP), and Compressive Sampling (CS) Methods

| | | | | |
|---|---|---|---|---|
| | 1.* | $1.49 \times 10^{-2}$ | $1.10 \times 10^{-3}$ | $1.03 \times 10^{-3}$ |
| | 0.9 | $1.02 \times 10^{-2}$ | $4.34 \times 10^{-3}$ | $2.58 \times 10^{-3}$ |
| | 0.8 | $1.04 \times 10^{-2}$ | $6.20 \times 10^{-3}$ | $4.24 \times 10^{-3}$ |
| **2000** | 0.7 | $1.14 \times 10^{-2}$ | $8.95 \times 10^{-3}$ | $6.93 \times 10^{-3}$ |
| | 0.6 | $1.40 \times 10^{-2}$ | $1.32 \times 10^{-2}$ | $1.10 \times 10^{-2}$ |
| | 0.5 | $1.76 \times 10^{-2}$ | $1.69 \times 10^{-2}$ | $1.59 \times 10^{-2}$ |
| | 0.4 | $2.38 \times 10^{-2}$ | $1.97 \times 10^{-2}$ | $1.93 \times 10^{-2}$ |

**Table 2**. Performance comparison of Boxcar (BOX), Thin-Plate (TP), and Compressive Sampling (CS) Methods (contd.)

**Algorithm 1.** Compressive Sampling Reconstruction of CV-fBm field by using TwIST

1. Complete the four quadrants of the complex-valued data (and corresponding locations) to obtain $\breve{e}_{PSF}^{sub}$ and $\breve{x}_{PSF}^{sub}$.

2. Convert the data from 2D into 1D array.

3. Construct observation matrix $\Phi$ based on $\breve{x}_{PSF}^{sub}$, i.e., set the entry to one when there are samples at corresponding location, otherwise set to zero.

4. Construct sparsity matrix $\Psi$, i.e., a complete set of the DFT bases

5. Construct matrix $A = \Phi\Psi$ and choose a particular value of $\lambda$

6. Solve the problem using TwIST and obtain $\breve{E}_{PSF}^{*}$.

7. Calculate estimated CV-fBm by inverse Fourier transform $\hat{\breve{e}}_{PSF} = IDFT(\breve{E}_{PSF}^{*})$.

8. Take one quadrant of $\hat{\breve{e}}_{PSF}$ to obtain the solution $\hat{e}_{PSF}$